\documentclass[runningheads]{llncs}
\usepackage{graphicx}
\usepackage{comment}
\usepackage{amsmath,amssymb} 
\usepackage{color}
\usepackage{multicol}
\usepackage{tabularx, makecell, multirow} 
\usepackage{wrapfig}

\usepackage[pagebackref=true,breaklinks=true,letterpaper=true,colorlinks,bookmarks=false]{hyperref}

\begin{document}
\pagestyle{headings}
\mainmatter
\def\ECCVSubNumber{2106}  

\title{ReActNet: Towards Precise Binary Neural Network with Generalized Activation Functions}

\titlerunning{ReActNet}
%
\author{Zechun Liu\inst{1,2}\thanks{Work done while visiting CMU. $^{\dagger}$ Corresponding author.} \and
Zhiqiang Shen\inst{2}$^{\dagger}$ \and
Marios Savvides\inst{2}\and
Kwang-Ting Cheng\inst{1}
}
\authorrunning{Z. Liu et al.}
\institute{$^1$ Hong Kong University of Science and Technology, $^2$ Carnegie Mellon University
\email{zliubq@connect.ust.hk, \{zhiqians,marioss\}@andrew.cmu.edu, timcheng@ust.hk}}
\maketitle

\begin{abstract}
In this paper, we propose several ideas for enhancing a binary network to close its accuracy gap from real-valued networks without incurring any additional computational cost. We first construct a baseline network by modifying and binarizing a compact real-valued network with parameter-free shortcuts, bypassing all the intermediate convolutional layers including the downsampling layers. This baseline network strikes a good trade-off between accuracy and efficiency, achieving superior performance than most of existing binary networks at approximately half of the computational cost. Through extensive experiments and analysis, we observed that the performance of binary networks is sensitive to activation distribution variations. Based on this important observation, we propose to generalize the traditional Sign and PReLU functions, denoted as RSign and RPReLU for the respective generalized functions, to enable explicit learning of the distribution reshape and shift at near-zero extra cost. Lastly, we adopt a distributional loss to further enforce the binary network to learn similar output distributions as those of a real-valued network. We show that after incorporating all these ideas, the proposed ReActNet outperforms all the state-of-the-arts by a large margin. Specifically, it outperforms Real-to-Binary Net and MeliusNet29 by 4.0\% and 3.6\% respectively for the top-1 accuracy and also reduces the gap to its real-valued counterpart to within 3.0\% top-1 accuracy on ImageNet dataset. Code and models are available at: \url{https://github.com/liuzechun/ReActNet}.
\end{abstract}

\section{Introduction}
The 1-bit convolutional neural network (1-bit CNN, also known as binary neural network)~\cite{courbariaux2016binarized,rastegari2016xnor}, of which both weights and activations are binary, has been recognized as one of the most promising neural network compression methods for deploying models onto the resource-limited devices. 
It enjoys 32$\times$ memory compression ratio, and up to 58$\times$ practical computational reduction on CPU, as demonstrated in~\cite{rastegari2016xnor}. Moreover, with its pure logical computation (\textit{i.e.}, XNOR operations between binary weights and binary activations), 1-bit CNN is both highly energy-efficient for embedded devices~\cite{ding2019regularizing,zhang2019dabnn}, and possesses the potential of being directly deployed on next generation memristor-based hardware~\cite{memristor-computing}. 

Despite these attractive characteristics of 1-bit CNN, the severe accuracy degradation prevents it from being broadly deployed. For example, a representative binary network, XNOR-Net~\cite{rastegari2016xnor} only achieves 51.2\% accuracy on the ImageNet classification dataset, leaving a $\sim$ 18\% accuracy gap from the real-valued ResNet-18. Some preeminent binary networks~\cite{ding2019regularizing,wang2019ci-bcnn} show good performance on small datasets such as CIFAR10 and MNIST, but still encounter severe accuracy drop when applied to a large dataset such as ImageNet.

In this study, our motivation is to further close the performance gap between binary neural networks and real-valued networks on the challenging large-scale datasets.
We start with designing a high-performance baseline network. Inspired by the recent advances in real-valued compact neural network design, we choose MobileNetV1~\cite{howard2017mobilenets} structure as our binarization backbone, which we believe is of greater practical value than binarizing non-compact models. 
Following the insights highlighted in \cite{liu2018bi}, we adopt blocks with identity shortcuts which bypass 1-bit vanilla convolutions to replace the convolutions in MobileNetV1. Moreover, we propose to use a concatenation of two of such blocks to handle the channel number mismatch in the downsampling layers, as shown in Fig.~\ref{fig:baseline_network}(a). 
This baseline network design not only helps avoid real-valued convolutions in shortcuts, which effectively reduces the computation to near half of that needed in prevalent binary neural networks~\cite{rastegari2016xnor,liu2018bi,martinez2020training}, but also achieves a high top-1 accuracy of 61.1\% on ImageNet.

\begin{figure}[t]
\begin{minipage}[h]{.5\linewidth}
\centering
\includegraphics[width=0.9\linewidth]{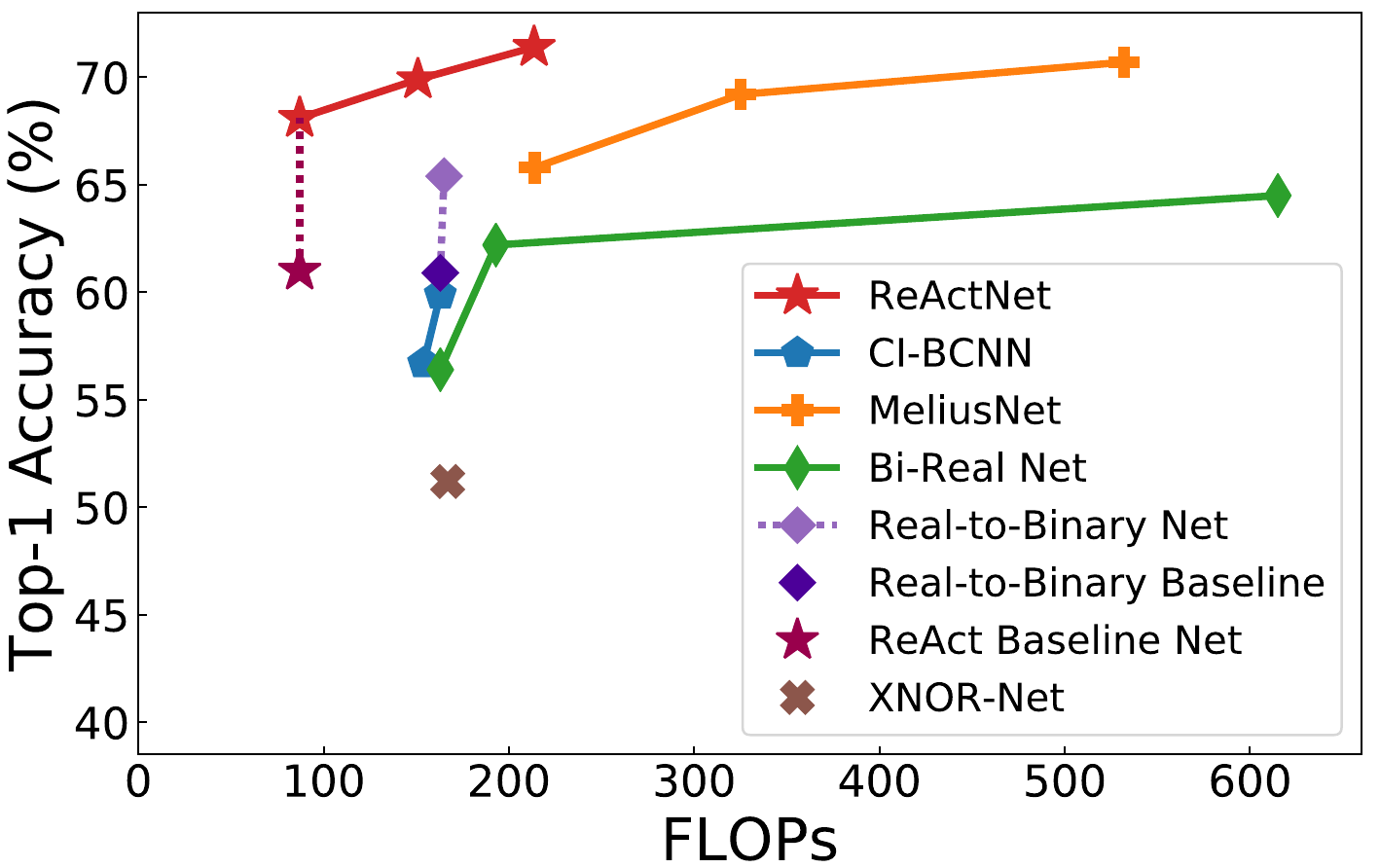}
\end{minipage}
\begin{minipage}[h]{.5\linewidth}
\centering
\resizebox{0.85\textwidth}{!}{
\begin{tabular}{ccccccccc}
\noalign{\smallskip}
\hline
\hline
Methods & OPs ($\times 10^8$) \ & \ Top-1 Acc(\%) \\ 
\hline
Real-to-Binary Baseline~\cite{martinez2020training} & 1.63 & 60.9\\
\textbf{Our ReAct Baseline Net} & \textbf{0.87} & \textbf{61.1}\\
\hline
XNORNet~\cite{rastegari2016xnor} & 1.67  & 51.2 \\
Bi-RealNet~\cite{liu2018bi} & 1.63  & 56.4 \\
Real-to-Binary~\cite{martinez2020training} & 1.65  & 65.4 \\
MeliusNet29~\cite{bethge2020meliusnet} & 2.14  & 65.8 \\
\textbf{Our ReActNet-A} & \textbf{0.87} & \textbf{69.4} \\ 
\hline
MeliusNet59~\cite{bethge2020meliusnet} & 5.32 & 70.7 \\
\textbf{Our ReActNet-C} & \textbf{2.14} & \textbf{71.4} \\ 
\hline
\hline
\end{tabular}}
\end{minipage}
\caption{\textbf{Computational cost vs. ImageNet Accuracy.} Proposed ReActNets significantly outperform other binary neural networks. In particular, ReActNet-C achieves state-of-the-art result with 71.4\% top-1 accuracy but being 2.5$\times$ more efficient than MeliusNet59. ReActNet-A exceeds Real-to-Binary Net and MeliusNet29 by 4.0\% and 3.6\% top-1 accuracy, respectively, and with more than 1.9$\times$ computational reduction. Details are described in Section~\ref{sec:sota}.}
\label{fig:pareto_chart}
\end{figure}

To further enhance the accuracy, we investigate another aspect which has not been studied in previous binarization or quantization works: activation distribution reshaping and shifting via non-linearity function design.
We observed that the overall activation value distribution affects the feature representation, and this effect will be exaggerated by the activation binarization. A small distribution value shift near zero will cause the binarized feature map to have a disparate appearance and in turn will influence the final accuracy. This observation will be elaborated in Section~\ref{sec:react_net}.
Enlightened by this observation, we propose a new generalization of Sign function and PReLU function to explicitly shift and reshape the activation distribution, denoted as ReAct-Sign (RSign) and ReAct-PReLU (RPReLU) respectively. These activation functions adaptively learn the parameters for distributional reshaping, which enhance the accuracy of the baseline network by $\sim$ 7\% with negligible extra computational cost. 

Furthermore, we propose a distributional loss to enforce the output distribution similarity between the binary and real-valued networks, which further boosts the accuracy by $\sim$ 1\%. After integrating all these ideas, the proposed network, dubbed as ReActNet, achieves 69.4\% top-1 accuracy on ImageNet with only 87M OPs, surpassing all previously published works on binary networks and reduce the accuracy gap from its real-valued counterpart to only 3.0\%, shown in Fig.~\ref{fig:pareto_chart}. 

We summarize our contributions as follows:

\begin{itemize}
    \item We design a baseline binary network by modifying MobileNetV1, whose performance already surpasses most of the previously published work on binary networks while incurring only half of the computational cost.
    \item We propose a simple channel-wise reshaping and shifting operation on the activation distribution, which helps binary convolutions spare the computational power in adjusting the distribution to learn more representative features.
    \item We further adopt a distributional loss between binary and real-valued network outputs, replacing the original loss, which facilitates the binary network to mimic the distribution of a real-valued network.
    \item We demonstrate that our proposed ReActNet, which integrates the above mentioned contributions, achieves 69.4\% top-1 accuracy on ImageNet, for the first time, exceeding the benchmarking ResNet-level accuracy (69.3\%) while achieving more than 22$\times$ reduction in computational complexity. This result also outperforms the state-of-the-art binary network~\cite{martinez2020training} by 4.0\% top-1 accuracy while incurring only half the OPs\footnote{OPs is a sum of binary OPs and floating-point OPs, i.e., OPs = BOPs/64 + FLOPs.}. 
\end{itemize}

\section{Related Work}

There have been extensive studies on neural network compression and acceleration, including quantization~\cite{zhuang2018towards,zhang2018lq,zhou2016dorefa}, pruning~\cite{ding2019global,he2017channel,liu2017learning,liu2019metapruning}, knowledge distillation~\cite{hinton2015distilling,shen2019meal,chen2017learning} and compact network design~\cite{howard2017mobilenets,sandler2018mobilenetv2,ma2018shufflenet,zhang2018shufflenet}. A comprehensive survey can be found in~\cite{sze2017efficient}. The proposed method falls into the category of quantization, specifically the extreme case of quantizing both weights and activations to only 1-bit, which is so-called network binarization or 1-bit CNNs. 

Neural network binarization originates from EBP~\cite{soudry2014expectation} and BNN~\cite{courbariaux2016binarized}, which establish an end-to-end gradient back-propagation framework for training the discrete binary weights and activations. As an initial attempt, BNN~\cite{courbariaux2016binarized} demonstrated its success on small classification datasets including CIFAR10~\cite{cifar10} and MNIST~\cite{mnist}, but encountered severe accuracy drop on a larger dataset such as ImageNet~\cite{imagenet}, only achieving 42.2\% top-1 accuracy compared to 69.3\% of the real-valued version of the ResNet-18. 

Many follow-up studies focused on enhancing the accuracy. XNOR-Net~\cite{rastegari2016xnor}, which proposed real-valued scaling factors to multiply with each of binary weight kernels, has become a representative binarization method and enhanced the top-1 accuracy to 51.2\%, narrowing the gap to the real-valued ResNet-18 to $\sim$18\%. Based on the XNOR-Net design, Bi-Real Net~\cite{liu2018bi} proposed to add shortcuts to propagate real-values along the feature maps, which further boost the top-1 accuracy to 56.4\%. 

Several recent studies attempted to improve the binary network performance via expanding the channel width~\cite{mishra2017wrpn}, increasing the network depth~\cite{liu2018bi_journal} or using multiple binary weight bases~\cite{lin2017abcnet}. Despite improvement to the final accuracy, the additional computational cost offsets the BNN’s high compression advantage.

For network compression, the real-valued network design used as the starting point for binarization should be compact. Therefore, we chose MobileNetV1 as the backbone network for development of our baseline binary network, which combined with several improvements in implementation achieves $\sim$ 2$\times$ further reduction in the computational cost compared to XNOR-Net and Bi-Real Net, and a top-1 accuracy of 61.1\%, as shown in Fig.~\ref{fig:pareto_chart}.

In addition to architectural design~\cite{bethge2020meliusnet,liu2018bi,phan2020binarizing}, studies on 1-bit CNNs expand from training algorithms~\cite{tang2017train,alizadeh2018empirical,zhuang2018towards,martinez2020training}, binary optimizer design~\cite{helwegen2019latent}, regularization loss design~\cite{ding2019regularizing,qin2020forward}, to better approximation of binary weights and activations~\cite{rastegari2016xnor,gu2019projection,wang2019ci-bcnn}. Different from these studies, this paper focuses on a new aspect that is seldom investigated before but surprisingly crucial for 1-bit CNNs’ accuracy, \textit{i.e.} activation distribution reshaping and shifting. For this aspect, we propose novel \textit{ReAct} operations, which are further combined with a proposed distributional loss. These enhancements improve the accuracy to 69.4\%, further shrinking the accuracy gap to its real-valued counterpart to only 3.0\%. The baseline network design and ReAct operations, as well as the proposed loss function are detailed in Section~\ref{sec:methodology}.

\section{Revisit: 1-bit Convolution}
In a 1-bit convolutional layer, both weights and activations are binarized to -1 and +1, such that the computationally heavy operations of floating-point matrix multiplication can be replaced by light-weighted bitwise XNOR operations and popcount operations~\cite{bulat2019xnor}, as:
\begin{align}
    \mathcal{X}_b * \mathcal{W}_b = {\rm popcount}({\rm XNOR}(\mathcal{X}_b,\mathcal{W}_b)),
\end{align}
where $\mathcal{W}_b$ and $\mathcal{X}_b$ indicate the matrices of binary weights and binary activations. Specifically, weights and activations are binarized through a sign function:
\begin{align}
{x_b = {\rm Sign}(x_r) = \left\{  
             \begin{array}{lr}  
             + 1, & {\rm if} \ x_r > 0 \\  
             - 1, & {\rm if} \ x_r \leq 0
             \end{array}  
\right. , \quad
w_b = \frac{||\mathcal{W}_r||_{l1}}{n}  {\rm Sign}(w_r) = \left\{  
             \begin{array}{lr}  
             + \frac{||\mathcal{W}_r||_{l1}}{n},  & {\rm if} \ w_r > 0 \\
             - \frac{||\mathcal{W}_r||_{l1}}{n},  & {\rm if} \ w_r \leq 0  
             \end{array} 
\right. 
}
\end{align}
The subscripts $b$ and $r$ denote binary and real-valued, respectively.
The weight binarization method is inherited from~\cite{rastegari2016xnor}, of which, $\frac{||\mathcal{W}_r||_{l1}}{n}$ is the average of absolute
weight values, used as a scaling factor to minimize the difference between binary and real-valued weights. XNOR-Net~\cite{rastegari2016xnor} also applied similar real-valued scaling factor to binary activations.
Note that with introduction of the proposed ReAct operations, to be described in Section~\ref{sec:react_net}, this scaling factor for activations becomes unnecessary and can be eliminated. 

\section{Methodology}
\label{sec:methodology}

In this section, we first introduce our proposed baseline network in Section~\ref{sec:baseline_network}. Then we analyze how the variation in activation distribution affects the feature quality and in turn influences the final performance. Based on this analysis, we introduce ReActNet which explicitly reshapes and shifts the activation distribution using ReAct-PReLU and ReAct-Sign functions described in Section~\ref{sec:react_net} and matches the outputs via a distributional loss defined between binary and real-valued networks detailed in Section~\ref{sec:kd}.

\subsection{Baseline Network}
\label{sec:baseline_network}
\begin{figure}[t]
\centering
\includegraphics[width=0.9\linewidth]{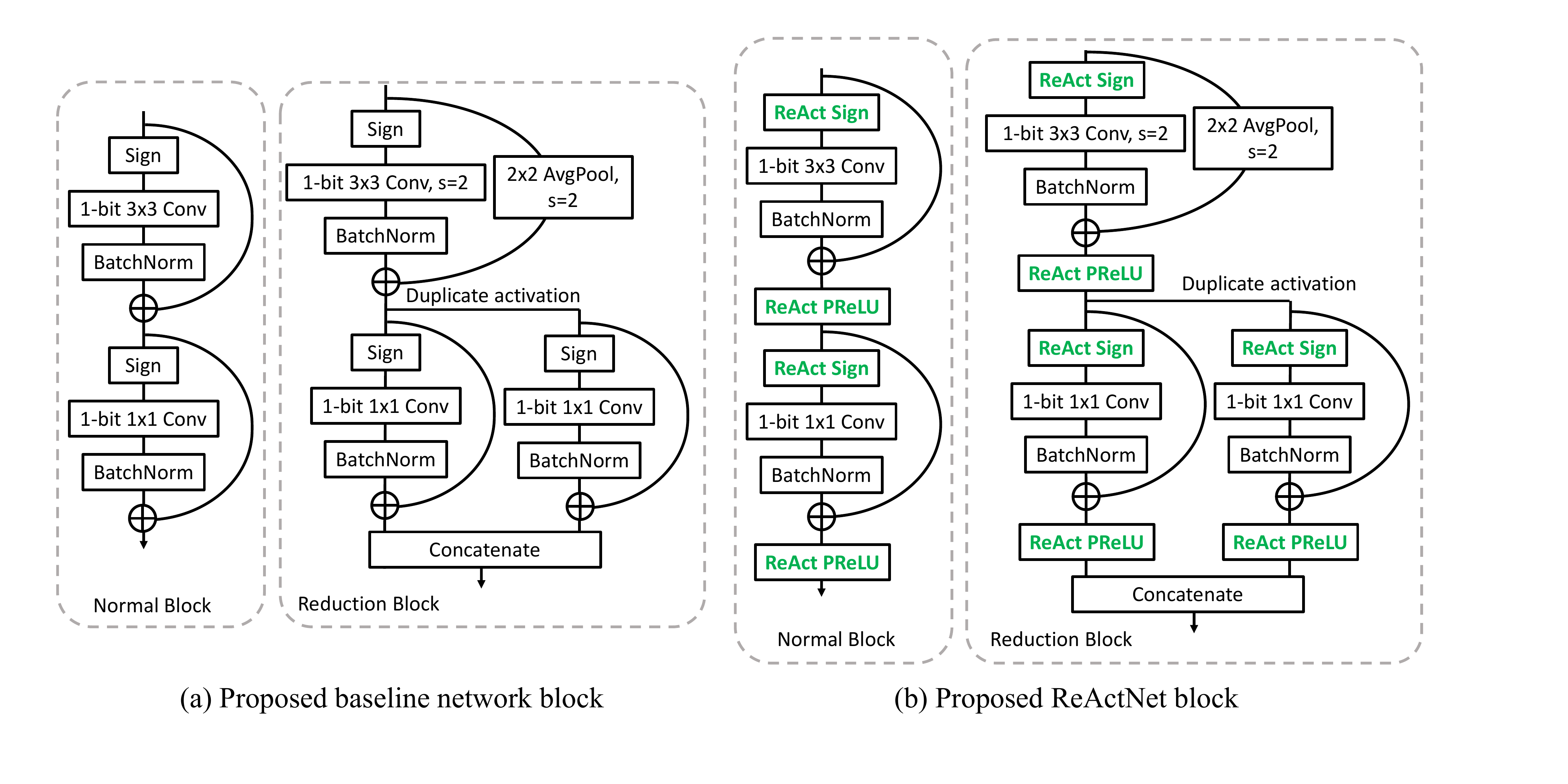}
\caption{The proposed baseline network modified from MobileNetV1~\cite{howard2017mobilenets}, which replaces the original (3$\times$3 depth-wise and 1$\times$1 point-wise) convolutional pairs by the proposed blocks. (a) The baseline’s configuration in terms of channel and layer numbers is identical to that of MobileNetV1. If the input and output channel numbers are equal in a dw-pw-conv pair in the original network, a normal block is used, otherwise a reduction block is adopted. For the reduction block, we duplicate the input activation and concatenate the outputs to increase the channel number. As a result, all 1-bit convolutions have the same input and output channel numbers and are bypassed by identity shortcuts. (b) In the proposed ReActNet block, ReAct-Sign and ReAct-PReLU are added to the baseline network. }
\label{fig:baseline_network}
\end{figure}

Most studies on binary neural networks have been binarizing the ResNet structure. However, further compressing compact networks, such as the MobileNets, would be more logical and of greater interest for practical applications. Thus, we chose MobileNetV1~\cite{howard2017mobilenets} structure for constructing our baseline binary network. 

Inspired by Bi-Real Net~\cite{liu2018bi}, we add a shortcut to bypass every 1-bit convolutional layer that has the same number of input and output channels. The 3$\times$3 depth-wise and the 1$\times$1 point-wise convolutional blocks in the MobileNetV1~\cite{howard2017mobilenets} are replaced by the 3$\times$3 and 1$\times$1 vanilla convolutions in parallel with shortcuts, respectively, as shown in Fig.~\ref{fig:baseline_network}. 

Moreover, we propose a new structure design to handle the downsampling layers. For the downsampling layers whose input and output feature map sizes differ, previous works~\cite{liu2018bi,wang2019ci-bcnn,martinez2020training} adopt real-valued convolutional layers to match their dimension and to make sure the real-valued feature map propagating along the shortcut will not be ``cut off'' by the activation binarization. However, this strategy increases the computational cost. Instead, our proposal is to make sure that all convolutional layers have the same input and output dimensions so that we can safely binarize them and use a simple identity shortcut for activation propagation without additional real-valued matrix multiplications. 

As shown in Fig.~\ref{fig:baseline_network}(a), we duplicate input channels and concatenate two blocks with the same inputs to address the channel number difference and also use average pooling in the shortcut to match spatial downsampling. All layers in our baseline network are binarized, except the first input convolutional layer and the output fully-connect layer. Such a structure is hardware friendly. 

\subsection{ReActNet}
\label{sec:react_net}
The intrinsic property of an image classification neural network is to learn a mapping from input images to the output logits. A logical deduction is that a good performing binary neural network should learn similar logits distribution as a real-valued network. However, the discrete values of variables limit binary neural networks from learning as rich distributional representations as real-valued ones. To address it, XNOR-Net~\cite{rastegari2016xnor} proposed to calculate analytical real-valued scaling factors and multiply them with the activations. Its follow-up works~\cite{xu2019accurate,bulat2019xnor} further proposed to learn these factors through back-propagation. 

In contrast to these previous works, this paper focuses on a different aspect: the activation distribution. We observed that small variations to activation distributions can greatly affect the semantic feature representations in 1-bit CNNs, which in turn will influence the final performance. However, 1-bit CNNs have limited capacity to learn appropriate activation distributions. To address this dilemma, we introduce generalized activation functions with learnable coefficients to increase the flexibility of 1-bit CNNs for learning semantically-optimized distributions.

\begin{figure}[t]
\centering
\includegraphics[width=0.85\linewidth]{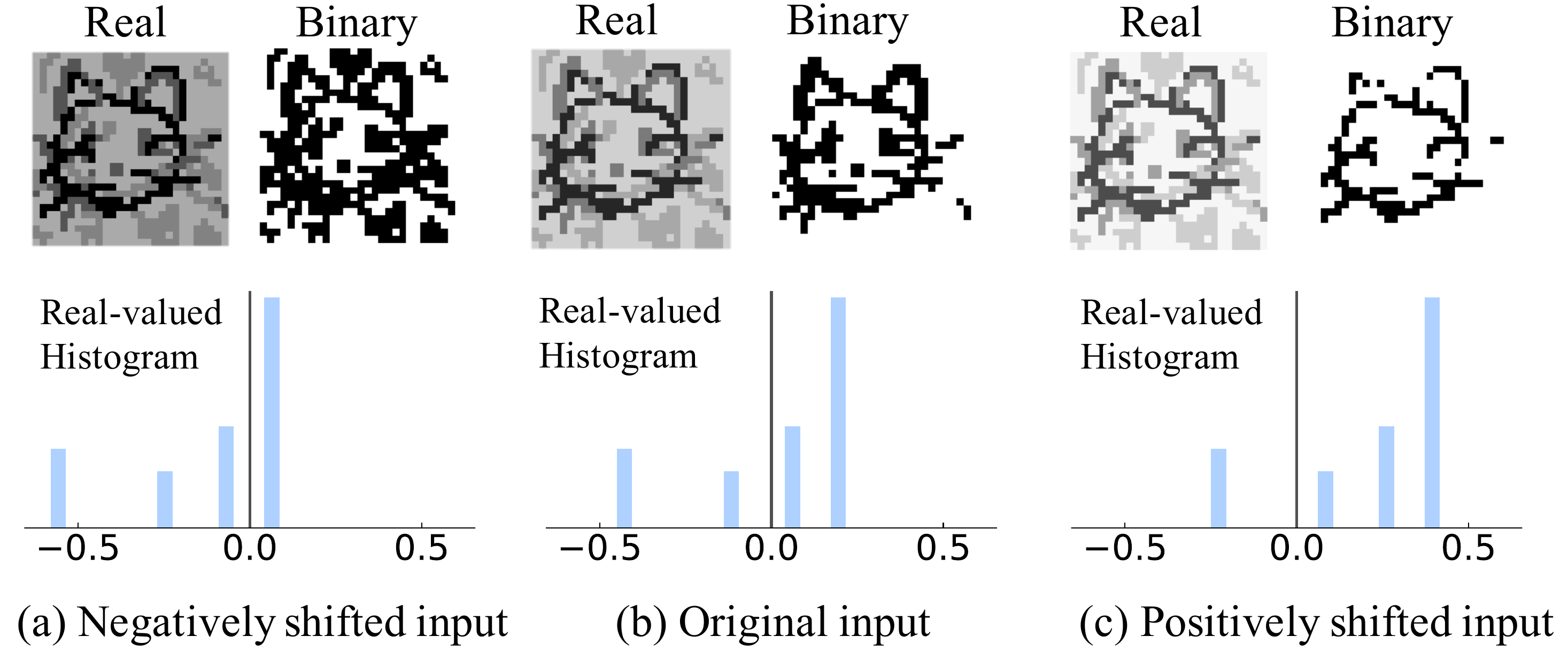}
\caption{An illustration of how distribution shift affects feature learning in binary neural networks. An ill-shifted distribution will introduce (a) too much background noise or (c) too few useful features, which harms feature learning.}
\label{fig:sign_distribution}
\end{figure}

\noindent{\textbf{Distribution Matters in 1-bit CNNs}} 
The importance of distribution has not been investigated much in training a real-valued network, because with weights and activations being continuous real values, reshaping or moving distributions would be effortless. 

However, for 1-bit CNNs, learning distribution is both crucial and difficult. Because the activations in a binary convolution can only choose values from $\{-1,+1\}$, making a small distributional shift in the input real-valued feature map before the sign function can possibly result in a completely different output binary activations, which will directly affect the informativeness in the feature and significantly impact the final accuracy. For illustration, we plot the output binary feature maps of real-valued inputs with the original (Fig.~\ref{fig:sign_distribution}(b)), positively-shifted (Fig.~\ref{fig:sign_distribution}(a)), and negatively-shifted (Fig.~\ref{fig:sign_distribution}(c)) activation distributions. Real-valued feature maps are robust to the shifts with which the legibility of semantic information will pretty much be maintained, while binary feature maps are sensitive to these shifts as illustrated in Fig.~\ref{fig:sign_distribution}(a) and Fig.~\ref{fig:sign_distribution}(c). 

\noindent{\textbf{Explicit Distribution Reshape and Shift via Generalized Activation Functions}}
Based on the aforementioned observation, we propose a simple yet effective operation to explicitly reshape and shift the activation distributions, dubbed as ReAct, which generalizes the traditional Sign and PReLU functions to ReAct-Sign (abbreviated as RSign) and ReAct-PReLU (abbreviated as RPReLU) respectively.

\noindent\textit{\textbf{Definition}}

\begin{figure}[t]
\centering
\resizebox{0.95\textwidth}{!}{
\includegraphics[width=1\linewidth]{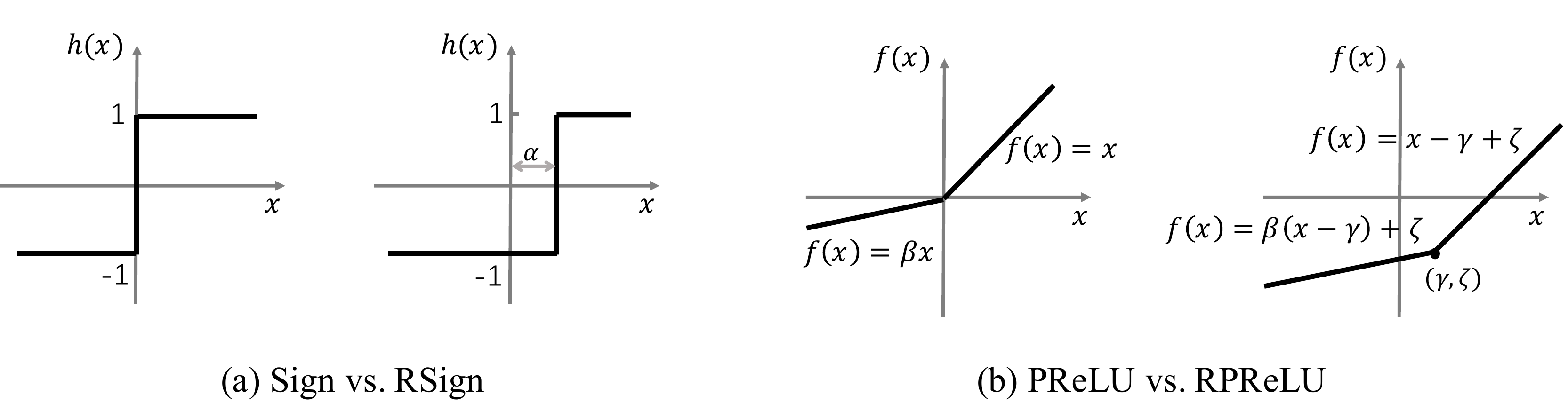}}
\caption{Proposed activation functions, RSign and RPReLU, with learnable coefficients and the traditional activation functions, Sign and PReLU.}
\label{fig:react_functions}
\end{figure}

Essentially, RSign is defined as a sign function with channel-wisely learnable thresholds: 
\begin{align}
x^b_i = h(x^r_i) = \left\{  
             \begin{array}{lr}  
             + 1, & \ \ {\rm if} \ x^r_i > \alpha_i \\  
             - 1, & \ \ {\rm if} \ x^r_i \leq \alpha_i \\
             \end{array}  
\right. 
.
\label{eq:RSign}
\end{align}
Here, $x^r_i$ is real-valued input of the RSign function $h$ on the \textit{i}th channel, $x^b_i$ is the binary output and $\alpha_i$ is a learnable coefficient controlling the threshold. The subscript \textit{i} in $\alpha_i$ indicates that the threshold can vary for different channels. The superscripts \textit{b} and \textit{r} refer to binary and real values. Fig.~\ref{fig:react_functions}(a) shows the shapes of RSign and Sign.

Similarly, RPReLU is defined as
\begin{align}
f(x_i) = \left\{  
             \begin{array}{lr}  
             x_i - \gamma_i + \zeta_i, & \ \ {\rm if} \ x_i > \gamma_i  \\  
             \beta_i (x_i  - \gamma_i) + \zeta_i, & \ \ {\rm if} \ x_i \leq \gamma_i
             \end{array} 
\right. 
,
\label{eq:RPReLU}
\end{align}
where $x_i$ is the input of the RPReLU function $f$ on the \textit{i}th channel, $\gamma_i$ and $\zeta_i$ are learnable shifts for moving the distribution, and $\beta_i$ is a learnable coefficient controlling the slope of the negative part. All the coefficients are allowed to be different across channels. Fig.~\ref{fig:react_functions}(b) compares the shapes of RPReLU and PReLU.

\begin{figure}[t]
\centering
\includegraphics[width=0.9\linewidth]{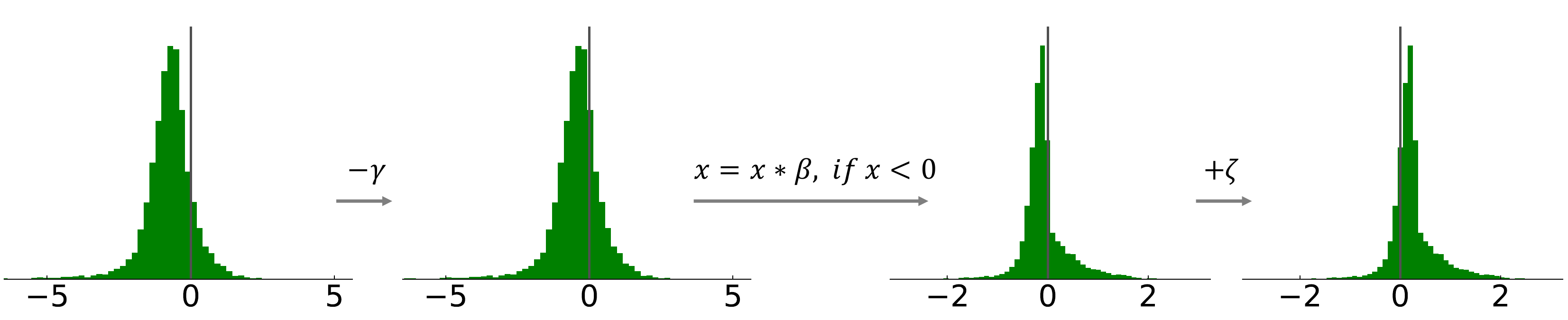}
\caption{An explanation of how proposed RPReLU operates. It first moves the input distribution by $-\gamma$, then reshapes the negative part by multiplying it with $\beta$ and lastly moves the output distribution by $\zeta$. }
\label{fig:rprelu}
\end{figure}

Intrinsically, RSign is learning the best channel-wise threshold ($\alpha$) for binarizing the input feature map, or equivalently, shifting the input distribution to obtain the best distribution for taking a sign. From the latter angle, RPReLU can be easily interpreted as $\gamma$ shifts the input distribution, finding a best point to use $\beta$ to ``fold'' the distribution, then $\zeta$ shifts the output distribution, as illustrated in Fig.~\ref{fig:rprelu}. These learned coefficients automatically adjust activation distributions for obtaining good binary features, which enhances the 1-bit CNNs' performance. With the introduction of these functions, the aforementioned difficulty in distributional learning can be greatly alleviated, and the 1-bit convolutions can effectively focus on learning more meaningful patterns. We will show later in the result section that this enhancement can boost the baseline network’s top-1 accuracy substantially.

The number of extra parameters introduced by RSign and RPReLU is only $ 4 \times number \ of \ channels $ in the network, which is negligible considering the large size of the weight matrices. The computational overhead approximates a typical non-linear layer, which is also trivial compared to the computational intensive convolutional operations. 

\noindent\textit{\textbf{Optimization}}

Parameters in RSign and RPReLU can be optimized end-to-end with other parameters in the network. The gradient of $\alpha_i$ in RSign can be simply derived by the chain rule as:
\begin{align}
    \frac{\partial \mathcal{L}}{\partial \alpha_i} = \sum_{x_i^r} \frac{\partial \mathcal{L}}{\partial h(x^r_i)} \frac{\partial h(x^r_i)}{\partial \alpha_i}, 
\end{align}
where $\mathcal{L}$ represents the loss function and $\frac{\partial \mathcal{L}}{\partial h(x^r_i)}$ denotes the gradients from deeper layers. The summation $\sum_{x_i^r}$ is applied to all entries in the \textit{i}th channel. The derivative $\frac{\partial h(x^r_i)}{\partial \alpha_i}$ can be easily computed as 
\begin{align}
    \frac{\partial h(x^r_i)}{\partial \alpha_i} = -1
\end{align}

Similarly, for each parameter in RPReLU, the gradients are computed with the following formula:
\begin{align}
    & \frac{\partial f(x_i)}{\partial \beta_i} = \textbf{I}_{\{ x_i \leq \gamma_i \}} \cdot (x-\gamma_i), \\
    & \frac{\partial f(x_i)}{\partial \gamma_i} = - \textbf{I}_{\{ x_i \leq \gamma_i \}} \cdot \beta_i -  \textbf{I}_{\{ x_i > \gamma_i \}},\\
    & \frac{\partial f(x_i)}{\partial \zeta_i} = 1.
\end{align}
Here, \textbf{I} denotes the indicator function. $\textbf{I}_{\{ \cdot \}}$ = 1 when the
inequation inside \{\} holds, otherwise $\textbf{I}_{\{ \cdot \}}$ = 0.

\subsection{Distributional Loss}
\label{sec:kd}
Based on the insight that if the binary neural networks can learn similar distributions as real-valued networks, the performance can be enhanced, we use a distributional loss to enforce this similarity, formulated as:
\begin{align}
    \mathcal{L}_{Distribution} = -\frac{1}{n}\sum_c\sum^n_{i=1} p_c^{\mathcal{R}_{\theta}}(X_i)\log(\frac{p_c^{\mathcal{B}_{\theta}}(X_i)}{p_c^{\mathcal{R}_{\theta}}(X_i)}),
\end{align}
where the distributional loss $\mathcal{L}_{Distribution}$ is defined as the KL divergence between the softmax output $p_c$ of a real-valued network $\mathcal{R}_{\theta}$ and a binary network $\mathcal{B}_{\theta}$. The subscript $c$ denotes classes and $n$ is the batch size.

Different from prior work~\cite{zhuang2018towards} that needs to match the outputs from every intermediate block, or further using multi-step progressive structural transition~\cite{martinez2020training}, we found that our distributional loss, while much simpler, can yield competitive results. Moreover, without block-wise constraints, our approach enjoys the flexibility in choosing the real-valued network without the requirement of architecture similarity between real and binary networks. 

\section{Experiments}
To investigate the performance of the proposed methods, we conduct experiments on ImageNet dataset. We first introduce the dataset and training strategy in Section~\ref{sec:exp_setting}, followed by comparison between the proposed networks and state-of-the-arts in terms of both accuracy and computational cost in Section~\ref{sec:sota}. We then analyze the effects of the distributional loss, concatenated downsampling layer and the RSign and the RPReLU in detail in the ablation study described in Section~\ref{sec:ablation}. Visualization results on how RSign and RPReLU help binary network capture the fine-grained underlying distribution are presented in Section~\ref{sec:visualization}. 

\subsection{Experimental Settings}
\label{sec:exp_setting}

\noindent\textbf{Dataset}
The experiments are carried out on the ILSVRC12 ImageNet classification dataset~\cite{imagenet}, which is more challenging than small datasets such as CIFAR~\cite{cifar10} and MNIST~\cite{mnist}. In our experiments, we use the classic data augmentation method described in~\cite{howard2017mobilenets}. 

\noindent\textbf{Training Strategy}
We followed the standard binarization method in~\cite{liu2018bi} and adopted the two-step training strategy as~\cite{martinez2020training}. In the first step, we train a network with binary activations and real-valued weights from scratch. In the second step, we inherit the weights from the first step as the initial value and fine-tune the network with weights and activations both being binary. For both steps, Adam optimizer with a linear learning rate decay scheduler is used, and the initial learning rate is set to 5e-4. We train it for 600k iterations with batch size being 256. The weight decay is set to 1e-5 for the first step and 0 for the second step.

\noindent\textbf{Distributional Loss}
In both steps, we use proposed distributional loss as the objective function for optimization, replacing the original cross-entropy loss between the binary network output and the label.

\noindent\textbf{OPs Calculation}
We follow the calculation method in~\cite{martinez2020training}, we count the binary operations (BOPs) and floating point operations (FLOPs) separately. The total operations (OPs) is calculated by OPs = BOPs/64 + FLOPs, following~\cite{rastegari2016xnor,liu2018bi}. 

\subsection{Comparison with State-of-the-art}
\label{sec:sota}

\begin{table}[t]
\centering
\resizebox{0.9\textwidth}{!}{
\begin{tabular}{ccc||ccccccc}
\noalign{\smallskip}
\hline
\hline
\multirowcell{2}{Methods} & \multirowcell{2}{Bitwidth \\ (W/A)} &  \multicolumn{1}{c||}{Acc(\%)} & \multirowcell{2}{Binary Methods} & \multirowcell{2}{BOPs \\ ($\times 10^9$)} & \multirowcell{2}{FLOPs \\ ($\times 10^8$)} & \multirowcell{2}{OPs \\ ($\times 10^8$)} &  \multicolumn{1}{c}{Acc(\%)} \\
& & Top-1 & & & & & Top-1\\
\hline
BWN~\cite{courbariaux2016binarized} & 1/32 & 60.8 & 
BNNs~\cite{courbariaux2016binarized} & 1.70 & 1.20 & 1.47 & 42.2 \\

TWN~\cite{li2016ternary} & 2/32 & 61.8 &
CI-BCNN~\cite{wang2019ci-bcnn}  & -- & -- & 1.63  & 59.9  \\

INQ~\cite{zhou2017incremental} & 2/32 & 66.0 &
Binary MobileNet~\cite{phan2020binarizing} & -- & -- & 1.54 & 60.9 \\

TTQ~\cite{zhu2016trained} & 2/32 & 66.6 &
PCNN~\cite{gu2019projection} & -- & -- & 1.63  & 57.3  \\

\cline{1-3}
SYQ~\cite{faraone2018syq} & 1/2 & 55.4 &
XNOR-Net~\cite{rastegari2016xnor} & 1.70 & 1.41 & 1.67  & 51.2 \\

HWGQ~\cite{cai2017deep} & 1/2 & 59.6 &
Trained Bin~\cite{xu2019accurate} & -- & -- & --& 54.2 \\

LQ-Nets~\cite{zhang2018lq} & 1/2 & 62.6 &
Bi-RealNet-18~\cite{liu2018bi} & 1.68 & 1.39 & 1.63  & 56.4 \\

DoReFa-Net~\cite{zhou2016dorefa} & 1/4 & 59.2 &
Bi-RealNet-34~\cite{liu2018bi} & 3.53 & 1.39 & 1.93  & 62.2 \\

\cline{1-3}
Ensemble BNN~\cite{zhu2019binary} & (1/1) $\times$ 6 & 61.1 &
Bi-RealNet-152~\cite{liu2018bi_journal} & 10.7 & 4.48 &  6.15  & 64.5 \\

Circulant CNN~\cite{liu2019circulant} & (1/1) $\times$ 4 & 61.4  & 
Real-to-Binary Net~\cite{martinez2020training} & 1.68 & 1.56  & 1.83  & 65.4 \\

Structured BNN~\cite{zhuang2019structured} & (1/1) $\times$ 4 & 64.2 &
MeliusNet29~\cite{bethge2020meliusnet} & 5.47 & 1.29 & 2.14  & 65.8 \\

Structured BNN*~\cite{zhuang2019structured} & (1/1) $\times$ 4 & 66.3 &
MeliusNet42~\cite{bethge2020meliusnet} & 9.69 & 1.74 & 3.25  & 69.2 \\

ABC-Net~\cite{lin2017abcnet} & (1/1) $\times$ 5 & 65.0 &
MeliusNet59~\cite{bethge2020meliusnet} & 18.3 & 2.45 & 5.32  & 70.7  \\

\hline
\textbf{Our ReActNet-A} & 1/1 & \multicolumn{1}{c}{--} & -- & \textbf{4.82} & \textbf{0.12} & \textbf{0.87} & \textbf{69.4} \\ 

\textbf{Our ReActNet-B} & 1/1 & \multicolumn{1}{c}{--} & -- & \textbf{4.69} & \textbf{0.44} & \textbf{1.63} & \textbf{70.1} \\ 

\textbf{Our ReActNet-C} & 1/1 & \multicolumn{1}{c}{--} & -- & \textbf{4.69} & \textbf{1.40} & \textbf{2.14} & \textbf{71.4} \\ 
\hline 
\hline
\noalign{\smallskip}
\noalign{\smallskip}
\end{tabular}}
\caption{Comparison of the top-1 accuracy with state-of-the-art methods. The left part presents quantization methods applied on ResNet-18 structure and the right part are binarization methods with varied structures (ResNet-18 if not specified). Quantization methods include weight quantization (upper left block), low-bit weight and activation quantization (middle left block) and the weight and activation binarization with the expanded network capacity (lower left block), where the number times (1/1) indicates the multiplicative factor. (W/A) represents the number of bits used in weight or activation quantization.}
\label{table:sota}
\end{table}

We compare ReActNet with state-of-the-art quantization and binarization methods. Table~\ref{table:sota} shows that ReActNet-A already outperforms all the quantizing methods in the left part, and also archives 4.0\% higher accuracy than the state-of-the-art Real-to-Binary Network~\cite{martinez2020training} with only approximately half of the OPs. Moreover, in contrast to~\cite{martinez2020training} which computes channel re-scaling for each block with real-valued fully-connected layers, ReActNet-A has pure 1-bit convolutions except the first and the last layers, which is more hardware-friendly.

To make further comparison with previous approaches that use real-valued convolution to enhance binary network’s accuracy~\cite{liu2018bi,martinez2020training,bethge2020meliusnet}, we constructed Re-ActNet-B and ReActNet-C, which replace the 1-bit 1$\times$1 convolution with real-valued 1$\times$1 convolution in the downsampling layers, as shown in Fig.~\ref{fig:downsampling_variation}(c). ReActNet-B defines the real-valued convolutions to be group convolutions with 4 groups, while ReActNet-C uses full real-valued convolution. We show that ReActNet-B achieves 13.7\% higher accuracy than Bi-RealNet-18 with the same number of OPs and ReActNet-C outperforms MeliusNet59 by 0.7\% with less than half of the OPs. 

Moreover, we applied the ReAct operations to Bi-RealNet-18, and obtained 65.5\% Top-1 accuracy, increasing the accuracy of Bi-RealNet-18 by 9.1\% without changing the network structure.

Considering the challenges in previous attempts to enhance 1-bit CNNs’ performance, the accuracy leap achieved by ReActNets is significant. It requires an ingenious use of binary networks’ special property to effectively utilize every precious bit and strike a delicate balance between binary and real-valued information. For example, ReActNet-A, with 69.4\% top-1 accuracy at 87M OPs, outperforms the real-valued 0.5$\times$ MobileNetV1 by 5.7\% greater accuracy at 41.6\% fewer OPs. These results demonstrate the potential of 1-bit CNNs and the effectiveness of our ReActNet design. 

\subsection{Ablation Study}
\label{sec:ablation}

\begin{figure}[t]
\centering
\includegraphics[width=\linewidth]{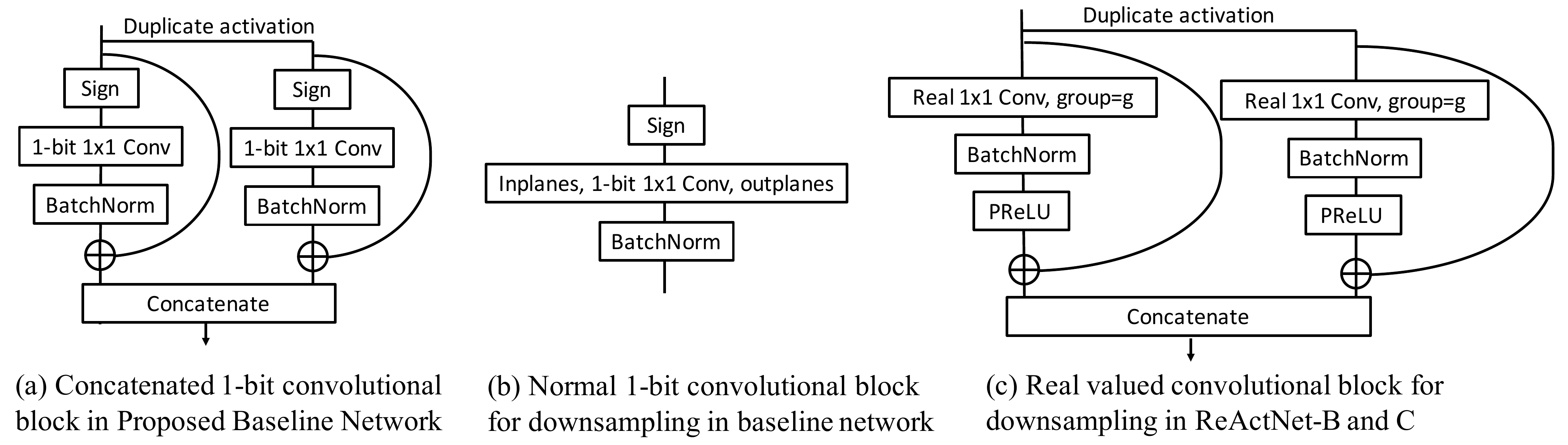}
\caption{Variations in the downsampling layer design.}
\label{fig:downsampling_variation}
\end{figure}

\begin{table}[t]
\centering
\resizebox{0.58\textwidth}{!}{
\begin{tabular}{lcccccc}
\noalign{\smallskip}
\hline
\hline
Network &  Top-1 Acc(\%) \\
\hline
Baseline network $\dagger$ * & 58.2\\
Baseline network $\dagger$ & 59.6\\
Proposed baseline network * & 61.1\\
Proposed baseline network & 62.5\\
\hline
Proposed baseline network + PReLU & 65.5\\
Proposed baseline network + RSign & 66.1\\
Proposed baseline network + RPReLU & 67.4\\
ReActNet-A (RSign and RPReLU) & 69.4 \\
\hline
Corresponding real-valued network & 72.4\\
\hline
\hline
\noalign{\smallskip}
\noalign{\smallskip}
\end{tabular}}
\caption{The effects of different components in ReActNet on the final accuracy. ($\dagger$ denotes the network not using the concatenated blocks, but directly binarizing the downsampling layers instead. * indicates not using the proposed distributional loss during training.)}
\label{table:ablation}
\end{table}
We conduct ablation studies to analyze the individual effect of the following proposed techniques: 

\noindent \textbf{Block Duplication and Concatenation}
Real-valued shortcuts are crucial for binary neural network accuracy~\cite{liu2018bi}. However, the input channels of the downsampling layers are twice the output channels, which violates the requirement for adding the shortcuts that demands an equal number of input and output channels. In the proposed baseline network, we duplicate the downsampling blocks and concatenate the outputs (Fig.~\ref{fig:downsampling_variation}(a)), enabling the use of identity shortcuts to bypass 1-bit convolutions in the downsampling layers. This idea alone results in a 2.9\% accuracy enhancement compared to the network without concatenation (Fig.~\ref{fig:downsampling_variation}(b)). The enhancement can be observed by comparing the 2\textit{nd} and 4\textit{th} rows of Table~\ref{table:ablation}.
With the proposed downsampling layer design, our baseline network achieves both high accuracy and high compression ratio. Because it no longer requires real-valued matrix multiplications in the downsampling layers, the computational cost is greatly reduced.
As a result, even without using the distributional loss in training, our proposed baseline network has already surpassed the Strong Baseline in~\cite{martinez2020training} by 0.1\% for top-1 accuracy at only half of the OPs. With this strong performance, this simple baseline network serves well as a new baseline for future studies on compact binary neural networks.

\noindent \textbf{Distributional Loss}
The results in the first section of Table~\ref{table:ablation} also validate that the distributional loss designed for matching the output distribution between binary and real-valued neural networks is effective for enhancing the performance of propose baseline network, improving the accuracy by 1.4\%, which is achieved independent of the network architecture design. 

\begin{figure}[t]
\centering
\includegraphics[width=0.9\linewidth]{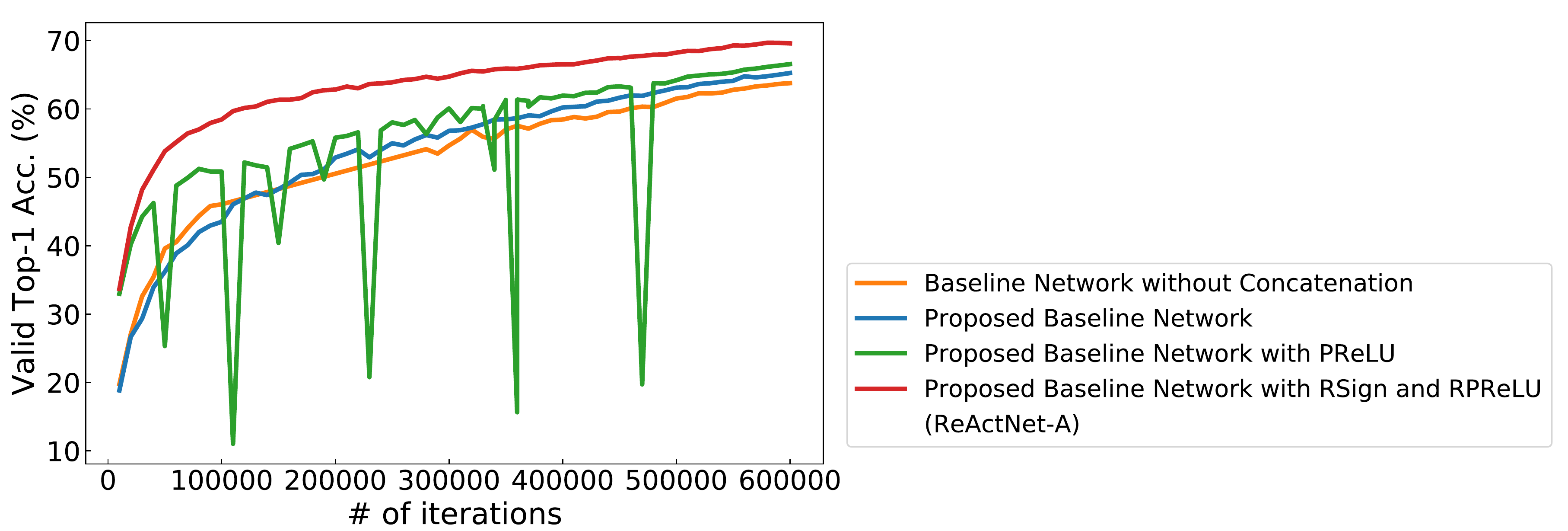}
\caption{Comparing validation accuracy curves between baseline networks and ReActNet. Using proposed RSign and RPReLU (red curve) achieves the higher accuracy and is more robust than using Sign and PReLU (green curve).}
\label{fig:acc_curve_comparison}
\end{figure}

\noindent \textbf{ReAct Operations}
The introduction of RSign and RPReLU improves the accuracy by 4.9\% and 3.6\% respectively over the proposed baseline network, as shown in the second section of Table~\ref{table:ablation}. By adding both RSign and RPReLU, ReActNet-A achieves 6.9\% higher accuracy than the baseline, narrowing the accuracy gap to the corresponding real-valued network to within 3.0\%. Compared to merely using the Sign and PReLU, the use of the generalized activation functions, RSign and RPReLU, with simple learnable parameters boost the accuracy by 3.9\%, which is very significant for the ImageNet classification task. 
As shown in Fig.~\ref{fig:acc_curve_comparison}, the validation curve of the network using original Sign + PReLU oscillates vigorously, which is suspected to be triggered by the slope coefficient $\beta$ in PReLU changing its sign which in turn affects the later layers with an avalanche effect. This also indirectly confirms our assumption that 1-bit CNNs are vulnerable to distributional changing. In comparison, the proposed RSign and RPReLU functions are effective for stabilizing training in addition to improving the accuracy. 

\subsection{Visualization}
\label{sec:visualization}
To help gain better insights, we visualize the learned coefficients as well as the intermediate activation distributions.

\begin{figure}[t]
\centering
\includegraphics[width=\linewidth]{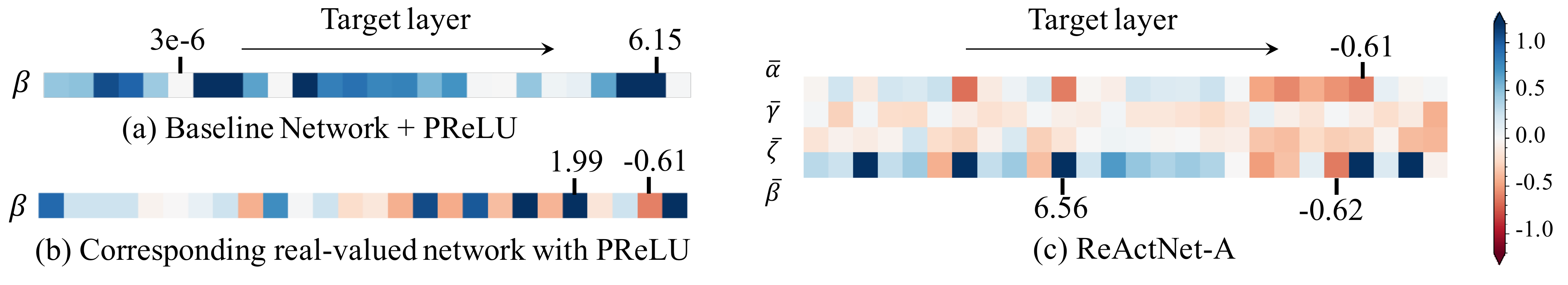}
\caption{The color bar of the learned coefficients. Blue color denotes the positive values while red denotes the negative, and the darkness in color reflects the absolute value. We also mark coefficients that have extreme values.}
\label{fig:react_param}
\end{figure}

\noindent\textbf{Learned Coefficients}
For clarity, we present the learned coefficients of each layer in form of the color bar in Fig.~\ref{fig:react_param}. Compared to the binary network using traditional PReLU whose learned slopes $\beta$ are positive only (Fig.~\ref{fig:react_param}(a)), ReActNet using RPReLU learns both positive and negative slopes (Fig.~\ref{fig:react_param}(c)), which are closer to the distributions of PReLU coefficients in a real-valued network we trained (Fig.~\ref{fig:react_param}(b)). The learned distribution shifting coefficients also have large absolute values as shown in Rows 1-3 of Fig.~\ref{fig:react_param}(c), indicating the necessity of their explicit shift for high-performance 1-bit CNNs. 

\begin{figure}[t]
\centering
\includegraphics[width=0.88\linewidth]{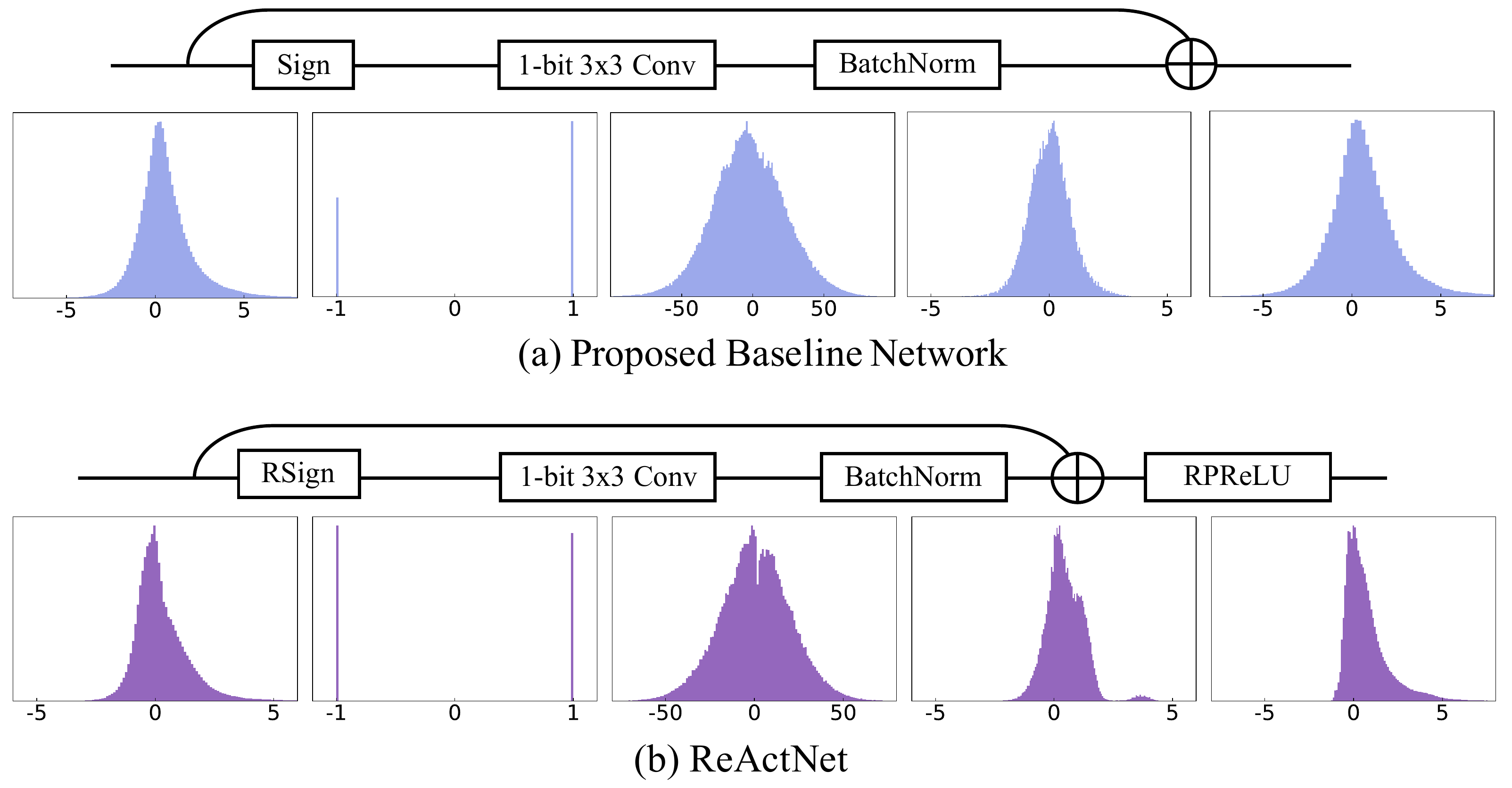}
\caption{Histogram of the activation distribution}
\label{fig:activation_distribution}
\end{figure}

\noindent\textbf{Activation Distribution}
In Fig.~\ref{fig:activation_distribution}, we show the histograms of activation distributions inside the trained baseline network and ReActNet. Compared to the baseline network without RSign and RPReLU, ReActNet’s distributions are more enriched and subtle, as shown in the forth sub-figure in Fig.~\ref{fig:activation_distribution}(b). Also, in ReActNet, the distribution of -1 and +1 after the sign function is more balanced, as illustrated in the second sub-figure in Fig.~\ref{fig:activation_distribution}(b), suggesting better utilization of black and white pixels in representing the binary features. 

\section{Conclusions}
In this paper, we present several new ideas to optimize a 1-bit CNN for higher accuracy. We first design parameter-free shortcuts based on MobileNetV1 to propagate real-valued feature maps in both normal convolutional layers as well as the downsampling layers. This yields a baseline binary network with 61.1\% top-1 accuracy at only 87M OPs for the ImageNet dataset. Then, based on our observation that 1-bit CNNs’ performance is highly sensitive to distributional variations, we propose ReAct-Sign and ReAct-PReLU to enable shift and reshape the distributions in a learnable fashion and demonstrate their dramatical enhancements on the top-1 accuracy. We also propose to incorporate a distributional loss, which is defined between the outputs of the binary network and the real-valued reference network, to replace the original cross-entropy loss for training. With contributions jointly achieved by these ideas, the proposed ReActNet achieves 69.4\% top-1 accuracy on ImageNet, which is just 3\% shy of its real-valued counterpart while at substantially lower computational cost.

%
%
\bibliographystyle{splncs04}
\bibliography{egbib}
\end{document}